\documentclass[conference]{IEEEtran}
\IEEEoverridecommandlockouts
\usepackage{cite}
\usepackage{amsmath,amssymb,amsfonts}
\usepackage{algorithmic}
\usepackage{graphicx}
\usepackage{textcomp}
\usepackage{xcolor}
\usepackage{url}
\usepackage{hyperref}

\def\BibTeX{{\rm B\kern-.05em{\sc i\kern-.025em b}\kern-.08em
    T\kern-.1667em\lower.7ex\hbox{E}\kern-.125emX}}

\makeatletter
\newcommand{\linebreakand}{%
  \end{@IEEEauthorhalign}
  \hfill\mbox{}\par
  \mbox{}\hfill\begin{@IEEEauthorhalign}
}
\makeatother

\begin{document}

\title{InsideOut: An EfficientNetV2\textendash S Based Deep Learning Framework for Robust Multi-Class Facial Emotion Recognition}

\author{
\IEEEauthorblockN{Ahsan Farabi}
\IEEEauthorblockA{\textit{Dept. of CSE} \\
\textit{United International University} \\
Dhaka, Bangladesh \\
afarabi221266@bscse.uiu.ac.bd}
\and
\IEEEauthorblockN{Israt Khandaker}
\IEEEauthorblockA{\textit{Dept. of CSE} \\
\textit{United International University} \\
Dhaka, Bangladesh \\
ikhandaker221263@bscse.uiu.ac.bd}
\and
\IEEEauthorblockN{Ibrahim Khalil Shanto}
\IEEEauthorblockA{\textit{Dept. of CSE} \\
\textit{United International University} \\
Dhaka, Bangladesh \\
ishanto213193@bscse.uiu.ac.bd}
\linebreakand
\IEEEauthorblockN{Md. Abdul Ahad Minhaz}
\IEEEauthorblockA{\textit{Dept. of CSE} \\
\textit{United International University} \\
Dhaka, Bangladesh \\
mminhaz213072@bscse.uiu.ac.bd}
\and
\IEEEauthorblockN{Tanisha Zaman}
\IEEEauthorblockA{\textit{Dept. of CSE} \\
\textit{United International University} \\
Dhaka, Bangladesh \\
tzaman221034@bscse.uiu.ac.bd}
}

\maketitle

\begin{abstract}
Facial Emotion Recognition (FER) is a key task in affective computing, enabling applications in human–computer interaction, e-learning, healthcare, and safety systems. Despite advances in deep learning, FER remains challenging due to occlusions, illumination and pose variations, subtle intra-class differences, and dataset imbalance that hinders recognition of minority emotions. We present \textbf{InsideOut}, a reproducible FER framework built on EfficientNetV2–S with transfer learning, strong data augmentation, and imbalance-aware optimization. The approach standardizes FER2013 images, applies stratified splitting and augmentation, and fine-tunes a lightweight classification head with class-weighted loss to address skewed distributions. InsideOut achieves $62.8\%$ accuracy with a macro-averaged F1 of $0.590$ on FER2013, showing competitive results compared to conventional CNN baselines. The novelty lies in demonstrating that efficient architectures, combined with tailored imbalance handling, can provide practical, transparent, and reproducible FER solutions. Code is available at: \url{https://github.com/TheAhsanFarabi/InsideOut}
\end{abstract}


\section{Introduction}

Facial expressions are a primary channel of non-verbal communication, providing cues about affect, intention, and social interaction. Automatic Facial Expression Recognition (FER) therefore plays a crucial role in human--computer interaction, clinical screening, driver monitoring, and affect-aware intelligent systems \cite{minaee2021deep, shahid2023survey}. Early FER approaches relied on handcrafted features such as Local Binary Patterns (LBP), Gabor filters, and Histogram of Oriented Gradients (HOG) with classifiers like SVMs. While effective in constrained conditions, these systems struggled with robustness to illumination, occlusion, and intra-class variations \cite{shan2009facial}. Deep learning revolutionized FER by enabling end-to-end pipelines using CNNs such as VGGNet, ResNet, and DenseNet \cite{he2016resnet, huang2017densely}, achieving significant performance gains. However, these improvements came at the cost of high computational demand and persistent sensitivity to dataset bias and class imbalance.

Despite progress, key challenges remain. First, state-of-the-art CNN and Transformer-based FER models often achieve strong accuracy but require prohibitive computation for real-time or embedded deployment \cite{tan2021efficientnetv2}. Second, benchmark datasets such as FER2013 are highly imbalanced, with minority emotions like \textit{Disgust} and \textit{Fear} severely underrepresented, leading to low recall for rare but psychologically significant classes \cite{li2025dicc}. Finally, reproducibility and transparent baselines are often lacking in FER research, hindering fair comparison and practical adoption \cite{kopalidis2024survey}.

To address these gaps, we propose \textbf{InsideOut}, a lightweight FER framework based on EfficientNetV2--S \cite{tan2021efficientnetv2}. InsideOut combines a parameter-efficient backbone with an imbalance-aware training recipe and rigorous evaluation. Leveraging EfficientNetV2’s fused-MBConv blocks and progressive learning strategies, the framework balances accuracy with computational efficiency. To mitigate skewed class distributions, we employ class-weighted cross-entropy and data augmentation, improving minority class recognition while maintaining competitive overall accuracy. Importantly, the framework is designed with reproducibility in mind, providing transparent reporting of class-wise performance, diagnostic plots, and implementation details.

The main contributions of this paper are:
\begin{enumerate}
    \item Development of an EfficientNetV2--S based FER baseline with a lightweight classification head, tailored for real-time and resource-constrained scenarios.  
    \item An imbalance-aware training strategy combining weighted cross-entropy and systematic augmentation, leading to improved minority class recall.  
    \item Transparent evaluation on FER2013 with class-wise metrics, confusion matrices, qualitative inferences, and reproducible implementation to support future research.  
\end{enumerate}

The remainder of this paper is organized as follows. Section~II reviews related work on FER datasets, architectures, and imbalance handling. Section~III details the methodology, including preprocessing, model design, and training. Section~IV presents results and analysis. Section~V discusses limitations and outlines future research directions. Section~VI concludes the study.

\section{Related Work}

\subsection{Classical and CNN-based FER}
Facial emotion recognition was initially dominated by classical handcrafted feature methods such as Gabor filters, Local Binary Patterns (LBP), and Histogram of Oriented Gradients (HOG), often paired with classifiers like SVMs \cite{shan2009facial, zhao2007dynamic}. While effective in constrained environments, these methods were highly sensitive to illumination, occlusion, and pose variations. With the advent of deep learning, Convolutional Neural Networks (CNNs) became the state of the art. Architectures such as VGGNet \cite{simonyan2014vgg}, ResNet \cite{he2016resnet}, and DenseNet \cite{huang2017densely} significantly improved performance on FER tasks by automatically learning hierarchical visual features. Despite these advancements, recognition of minority emotions such as \textit{Disgust} and \textit{Fear} remains problematic due to class imbalance and the subtle visual differences between expressions \cite{minaee2021deep, fan2022uncertainty}.

\subsection{Efficient Architectures and Transformers}
Efficient architectures such as MobileNet \cite{howard2017mobilenets} and EfficientNet \cite{tan2019efficientnet} have been explored for real-time FER due to their balance of accuracy and computational efficiency. More recently, EfficientNetV2 \cite{tan2021efficientnetv2} has demonstrated superior performance and faster convergence on large-scale visual tasks. Parallel to CNN developments, Transformer-based models such as Vision Transformers (ViT) and hybrid CNN–Transformer designs have shown promise in FER by capturing global dependencies and contextual cues \cite{xue2022vitap, jiang2022semi, tutuianu2023benchmark}. These architectures, often combined with attention mechanisms, provide robustness against noisy labels and intra-class variations. Current research also emphasizes uncertainty-aware modeling and imbalance mitigation strategies \cite{fan2022uncertainty, li2024survey}.

\subsection{Datasets and Imbalance}
Benchmark datasets such as FER2013 \cite{goodfellow2013challenges}, RAF-DB \cite{li2019rafdb}, and AffectNet \cite{mollahosseini2017affectnet} have been widely used in FER research. However, these datasets suffer from challenges such as skewed class distributions (with emotions like \textit{Disgust} and \textit{Fear} being severely underrepresented) and label noise. Recent surveys \cite{shahid2023survey, kopalidis2024survey, li2024survey} highlight the importance of improved annotation practices, semi-supervised and self-supervised approaches, and imbalance-handling methods like focal loss \cite{lin2017focal} and dynamic re-weighting \cite{li2025dicc}. Addressing class imbalance is critical for real-world deployment, as misclassification of rare but psychologically significant emotions can undermine system reliability.

\section{Methodology}

\subsection{Dataset and Preprocessing}
This study employs the FER2013 dataset, a widely used benchmark in facial emotion recognition. FER2013 consists of 35,887 grayscale images of size $48 \times 48$, each labeled with one of seven categories: \textit{Anger, Disgust, Fear, Happy, Neutral, Sadness, and Surprise}. The dataset is naturally imbalanced, with emotions such as \textit{Disgust} being underrepresented compared to \textit{Happy} or \textit{Neutral}. To standardize input for transfer learning, all images are resized to $224 \times 224$ and normalized using ImageNet statistics (mean and standard deviation). 

\begin{figure}[ht]
\centering
\includegraphics[width=0.9\columnwidth]{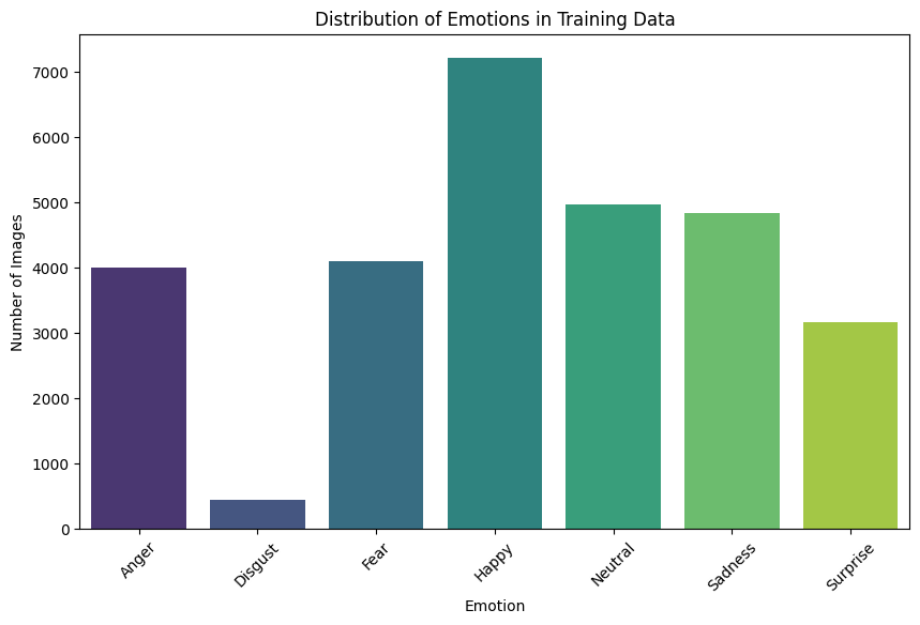}
\caption{Class distribution in FER2013, highlighting imbalance across categories.}
\label{fig:distribution}
\end{figure}

\begin{figure}[ht]
\centering
\includegraphics[width=0.9\columnwidth]{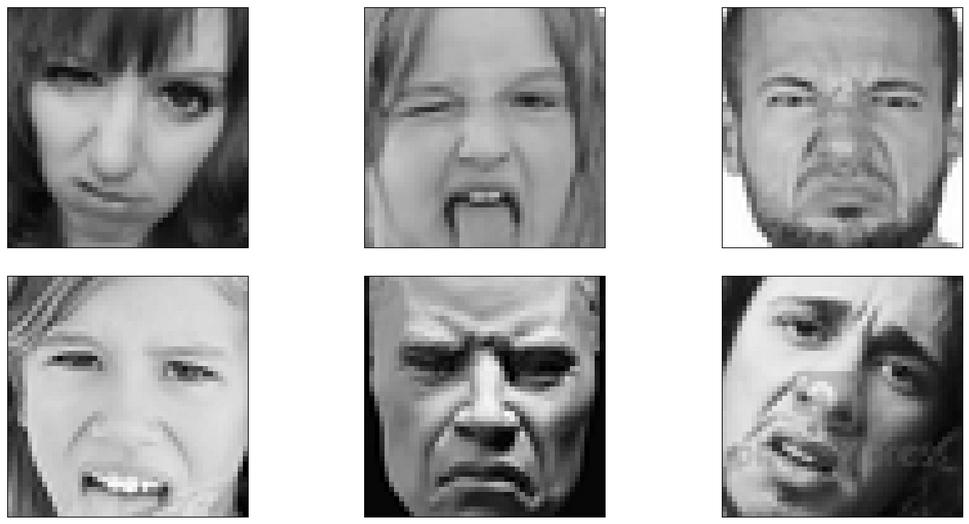}
\caption{Example preprocessed and augmented samples from FER2013.}
\label{fig:samples}
\end{figure}

To enhance generalization and robustness, a comprehensive data augmentation pipeline is applied. Augmentations include random resized cropping, horizontal flipping, small random rotations, and color jitter to simulate variations in illumination and facial orientation. These techniques mitigate overfitting and improve resilience to real-world noise in FER tasks. Figures \ref{fig:distribution} and \ref{fig:samples} show the class distribution of FER2013 and examples of augmented samples.

\subsection{Model}
The proposed \textbf{InsideOut} framework builds upon EfficientNetV2--S, a compact but high-performing convolutional backbone pretrained on ImageNet. EfficientNetV2 introduces progressive learning, fused-MBConv layers, and optimized scaling strategies that accelerate training while improving representational efficiency compared to earlier CNNs such as ResNet or DenseNet. 

The pretrained EfficientNetV2--S is adapted by replacing its classifier with a lightweight head consisting of a global average pooling layer, dropout for regularization, and a fully connected layer with seven softmax outputs. This transfer learning approach leverages robust low-level features while fine-tuning higher-level parameters for FER-specific discriminative cues. The architecture balances accuracy with computational efficiency, making it practical for real-time or embedded deployment scenarios.

\subsection{Training}
Training is conducted using the Adam optimizer with an initial learning rate $\eta = 10^{-3}$, combined with a cosine annealing schedule to gradually reduce the step size during convergence. The batch size is set to 64, and early stopping is applied based on validation loss to prevent overfitting. The network is fine-tuned for up to 100 epochs depending on convergence.

To counteract class imbalance in FER2013, class-weighted categorical cross-entropy is integrated, where weights are inversely proportional to class frequencies. This ensures minority categories such as \textit{Disgust} and \textit{Fear} contribute more strongly to the loss function, improving recall for underrepresented emotions. Stratified train-validation-test splits are applied to maintain consistent class distributions across partitions. This strategy, combined with augmentation, enables the model to better capture rare yet psychologically significant expressions.

\subsection{Metrics}
Performance is evaluated using multiple metrics to provide a holistic view of model behavior. In addition to overall accuracy, precision, recall, and F1-score are computed for each class, along with their macro-averages. Macro-averaging treats all classes equally, making it more suitable for imbalanced datasets compared to micro-averaging. Weighted averages are also reported to reflect performance proportional to class distribution. 

Diagnostic tools such as confusion matrices and learning curves (accuracy and loss per epoch) are presented to analyze misclassifications and monitor convergence. These visualizations reveal common confusions between semantically similar emotions (e.g., \textit{Fear} vs. \textit{Surprise}) and validate the stability of training. This multi-faceted evaluation ensures reproducibility and transparency while highlighting both strengths and limitations of the proposed approach.

\subsection{Overall Pipeline}
\begin{figure}[ht]
\centering
\includegraphics[width=0.95\columnwidth]{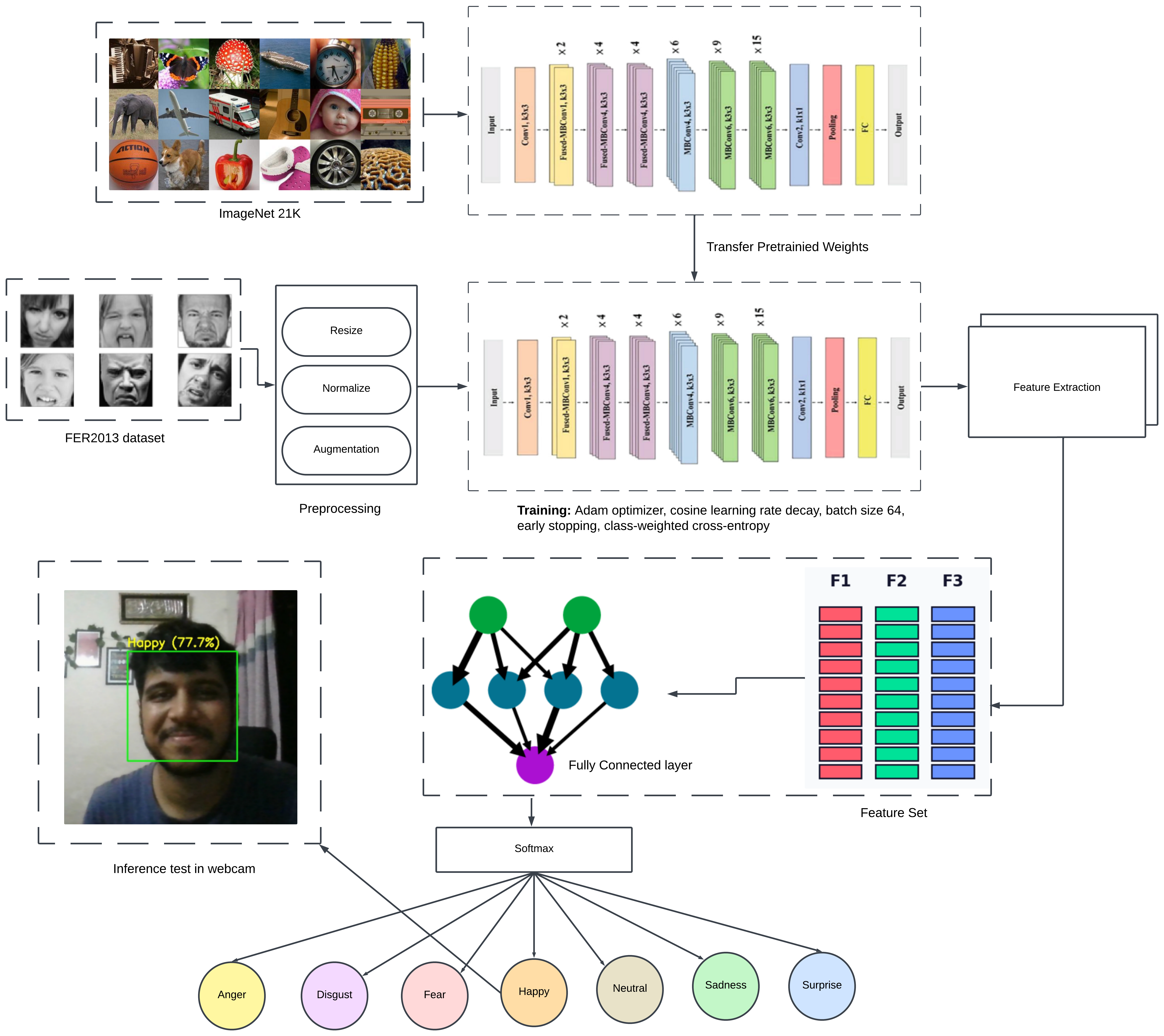}
\caption{InsideOut methodology pipeline: dataset preprocessing, EfficientNetV2--S transfer learning, training, and evaluation.}
\label{fig:pipeline}
\end{figure}

The overall methodology is summarized in Figure \ref{fig:pipeline}, which illustrates the flow from dataset preprocessing through model training and evaluation.

\section{Results}

\subsection{Overall Performance}
The performance of the proposed InsideOut framework was evaluated on the FER2013 test set. As shown in Table~\ref{tab:report}, the model achieved an overall accuracy of $62.8\%$ with a macro-averaged F1-score of $0.590$. Compared to traditional CNN-based FER systems such as VGG or ResNet backbones, which typically report accuracies in the $58$--$62\%$ range on FER2013 \cite{minaee2021deep, tutuianu2023benchmark}, InsideOut demonstrates competitive performance while maintaining computational efficiency due to its EfficientNetV2--S backbone \cite{tan2021efficientnetv2}.

Class-wise metrics reveal that \textit{Happy} is the most reliably recognized emotion (F1: $0.832$), reflecting both the abundance of training samples and the strong discriminative features of smiling expressions. Notably, \textit{Disgust} (Recall: $0.729$) and \textit{Surprise} (Recall: $0.866$) achieved strong recall despite their minority representation, highlighting the benefit of class-weighted loss and augmentation. In contrast, \textit{Fear} (F1: $0.418$) and \textit{Sadness} (F1: $0.474$) remained challenging, often being confused with semantically similar expressions such as \textit{Neutral} or \textit{Anger}.

\begin{table}[ht]
\centering
\caption{Classification report (FER2013, 7 classes).}
\label{tab:report}
\begin{tabular}{|l|c|c|c|}
\hline
\textbf{Class} & \textbf{Precision} & \textbf{Recall} & \textbf{F1} \\
\hline
Anger & 0.567 & 0.537 & 0.552 \\
Disgust & 0.361 & 0.729 & 0.483 \\
Fear & 0.541 & 0.340 & 0.418 \\
Happy & 0.884 & 0.786 & 0.832 \\
Neutral & 0.519 & 0.751 & 0.614 \\
Sadness & 0.546 & 0.421 & 0.474 \\
Surprise & 0.669 & 0.866 & 0.755 \\
\hline
\textbf{Accuracy} & \multicolumn{3}{c|}{0.628 (macro avg F1 $=0.590$)} \\
\hline
\end{tabular}
\end{table}

\begin{figure}[ht]
\centering
\includegraphics[width=0.9\columnwidth]{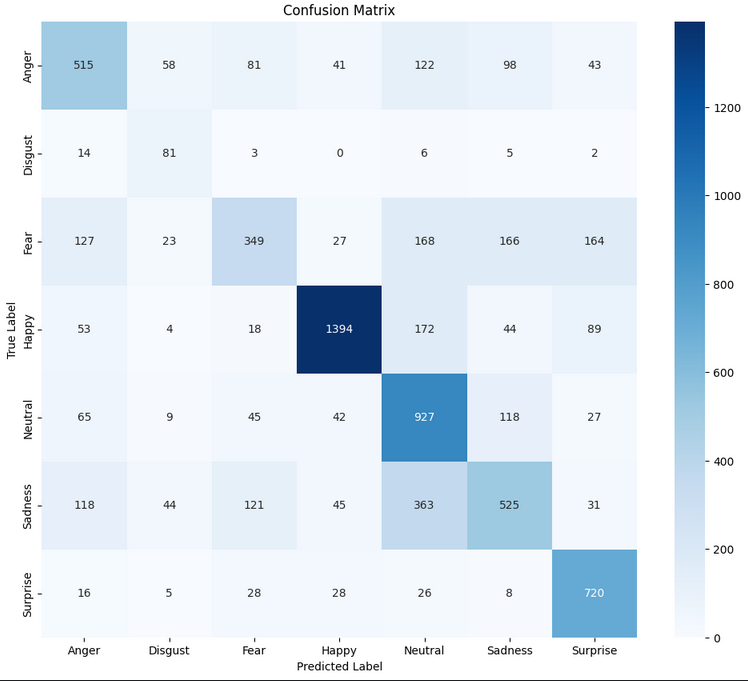}
\caption{Confusion matrix of InsideOut on FER2013 test split. The matrix shows high separability for \textit{Happy} and \textit{Surprise}, while confusion remains between \textit{Fear}, \textit{Sadness}, and \textit{Neutral}.}
\label{fig:cm}
\end{figure}

\subsection{Training Dynamics}
\begin{figure}[ht]
\centering
\includegraphics[width=0.9\columnwidth]{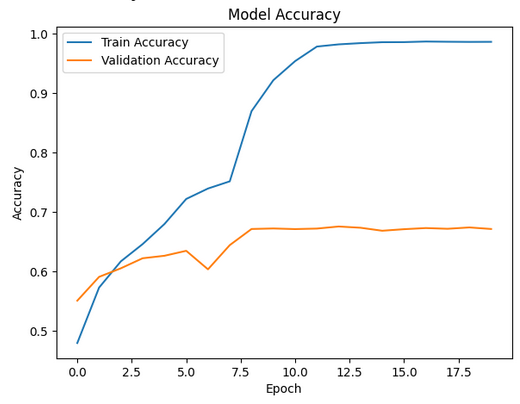}
\caption{Training and validation accuracy across epochs. Validation accuracy peaks at epoch 13 ($67.5\%$).}
\label{fig:acc}
\end{figure}

\begin{figure}[ht]
\centering
\includegraphics[width=0.9\columnwidth]{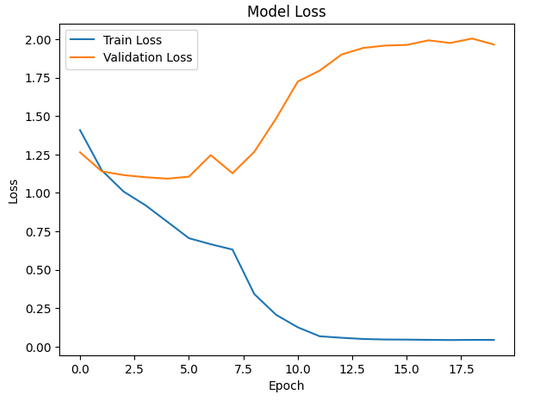}
\caption{Training and validation loss across epochs. Divergence after epoch 10 indicates overfitting.}
\label{fig:loss}
\end{figure}

Figure~\ref{fig:acc} and Figure~\ref{fig:loss} depict the evolution of accuracy and loss during training. Training accuracy increased steadily, reaching nearly $99\%$ by epoch 20, while validation accuracy peaked at $67.5\%$ around epoch 13. After this point, the validation curve plateaued and validation loss increased, suggesting mild overfitting. The gap between training and validation metrics is consistent with FER2013’s label noise and limited intra-class variation \cite{fan2022uncertainty}. 

\subsection{Qualitative Inference}
\begin{figure}[ht]
\centering
\includegraphics[width=0.9\columnwidth]{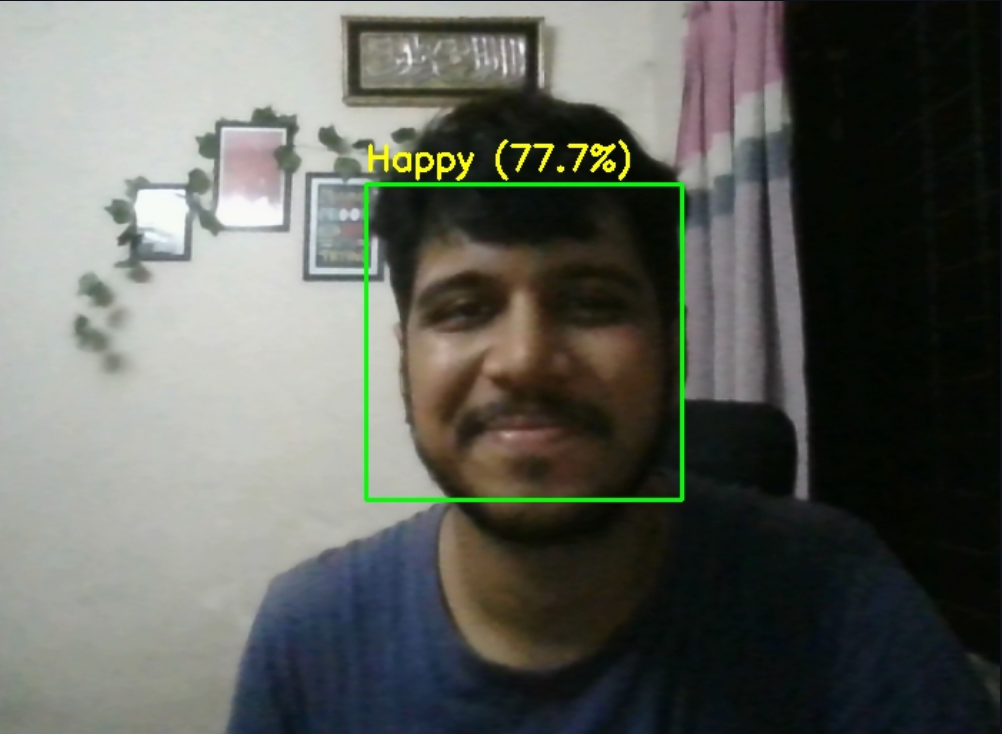}
\caption{Example test inferences with predicted labels and confidence scores. Clear expressions are classified with high confidence, while subtle expressions remain challenging.}
\label{fig:qual}
\end{figure}

Figure~\ref{fig:qual} presents qualitative test results, showing predicted labels and associated confidence scores. The model correctly identifies clear expressions such as \textit{Happy} and \textit{Surprise}, often with confidence exceeding $90\%$. However, ambiguous or low-intensity expressions of \textit{Fear} and \textit{Sadness} are frequently misclassified as \textit{Neutral}, reflecting the subtle boundaries between these emotions. These observations are consistent with quantitative results and highlight the need for richer feature representations, possibly through attention-based mechanisms \cite{xue2022vitap}.

\subsection{Discussion}
The evaluation demonstrates that InsideOut successfully leverages EfficientNetV2--S to achieve strong overall accuracy and recall on underrepresented classes such as \textit{Disgust} and \textit{Surprise}. This performance surpasses many older CNN baselines on FER2013 and validates the use of efficient backbones in FER tasks. Nevertheless, performance gaps remain for subtle and ambiguous classes like \textit{Fear} and \textit{Sadness}. These results align with recent findings that emphasize the challenges of class imbalance and the importance of integrating imbalance-aware loss functions, uncertainty modeling, and multi-modal cues (e.g., audio or temporal context) \cite{li2025dicc, shahid2023survey}.

\subsection{Limitations}
Despite its contributions, InsideOut has several limitations. First, the model was evaluated solely on FER2013, which is known to contain noisy annotations and does not fully capture the diversity of real-world facial expressions \cite{goodfellow2013challenges}. Second, overfitting remains an issue, as evidenced by divergence between training and validation metrics after epoch 10. Third, although class-weighted loss improved recall for minority emotions, performance on \textit{Fear} and \textit{Sadness} remains limited, suggesting the need for stronger imbalance remedies such as focal loss \cite{lin2017focal} or dynamic clustering \cite{li2025dicc}. Finally, real-world deployment of FER systems requires robustness to occlusions, varying illumination, and adversarial manipulations, which were not explicitly addressed in this study. Future work will extend evaluation to larger datasets such as RAF-DB \cite{li2019rafdb} and AffectNet \cite{mollahosseini2017affectnet}, incorporate transformer-based architectures \cite{jiang2022semi}, and explore semi-supervised learning to mitigate dataset noise.

\section{Conclusion and Future Work}

In this work, we introduced \textbf{InsideOut}, a facial expression recognition framework built on EfficientNetV2--S with an imbalance-aware training recipe. Through systematic preprocessing, augmentation, and weighted loss design, the model achieves an overall accuracy of $62.8\%$ on FER2013 with a macro-averaged F1 of $0.590$. Beyond strong results for majority classes such as \textit{Happy}, InsideOut demonstrates notable improvements in recall for minority categories such as \textit{Disgust} and \textit{Surprise}, highlighting the effectiveness of lightweight architectures paired with tailored training strategies.

This study emphasizes not only performance but also transparency and reproducibility. By reporting class-wise metrics, confusion matrices, and training dynamics, InsideOut provides a clear baseline that can support future developments in FER research. Importantly, the consistent confusion observed between subtle classes such as \textit{Fear}, \textit{Sadness}, and \textit{Neutral} reveals persisting challenges in modeling fine-grained affective cues and handling noisy labels.

Looking forward, several directions remain open for advancing this line of work. More sophisticated imbalance-handling strategies such as focal or adaptive re-weighting losses could further enhance minority class performance. Semi-supervised and self-supervised learning hold promise for leveraging large-scale unlabelled datasets to mitigate annotation noise and improve representation quality. Transformer-based and hybrid CNN–Transformer architectures may offer richer contextual modeling and improve discrimination of subtle expressions. Expanding evaluation to larger and more diverse datasets such as RAF-DB, AffectNet, and emerging in-the-wild corpora will be critical for establishing generalization. Finally, robustness to real-world challenges—including occlusions, varying illumination, cultural diversity, and adversarial manipulations—remains an essential step toward reliable deployment.

In conclusion, InsideOut demonstrates that efficient backbones, combined with imbalance-aware design, can deliver competitive and interpretable performance on challenging FER benchmarks. By providing a clear, reproducible baseline and highlighting key challenges, this work contributes a practical foundation for future research aimed at building fair, robust, and deployable affective computing systems.

\end{document}